# A natural language processing-based approach: mapping human perception by understanding deep semantic features in street view images

Haoran Ma[1] and Dongdong Wu[2]

[1] School of Design, Jiangnan University, Wuxi, China
(6210307146@stu.jiangnan.edu.cn)
2 School of Computer Science and Engineering, Southeast University, Nanjing, China
(dongdongwu@seu.edu.cn)

**Abstract.** In the past decade, using Street View images and machine learning to measure human perception has become a mainstream research approach in urban science. However, this approach using only image-shallow information makes it difficult to comprehensively understand the deep semantic features of human perception of a scene. In this study, we proposed a new framework based on a pre-train natural language model to understand the relationship between human perception and the sense of a scene. Firstly, Place Pulse 2.0 was used as our base dataset, which contains a variety of human-perceived labels, namely, beautiful, safe, wealthy, depressing, boring, and lively. And an image captioning network was used to extract the description information of each street view image. Secondly, a pre-trained BERT model was finetuning and added a regression function for six human perceptual dimensions. Furthermore, we compared the performance of five traditional regression methods with our approach and conducted a migration experiment in Hong Kong. Our results show that human perception scoring by deep semantic features performed better than previous studies by machine learning methods with shallow features. The use of deep scene semantic features provides new ideas for subsequent human perception research, as well as better explanatory power in the face of spatial heterogeneity.

**Keywords:** Natural language processing · Human perception · Street view images · Pretrained model

## 1 Introduction

Urban experience has always been captured, expressed, and disseminated through human language description or digital photography and recording. Previous research has demonstrated the importance of measuring human perception in comprehending urban spatial variability and elucidating the influence of urban function. There has long been a lot of interest in the research of effects on participants' scene perception (Lynch, 1962, Kaplan and Kaplan, 1989). Recently, related research in the area of urban science has successfully utilized the quickly evolving urban street view images (Zhang et al., 2018, Wang et al., 2022b).

These pieces, however, focus more on the relationship between environmental factors and urban street settings with limited information and are unable to adequately represent urban features from a linguistic perspective. As a result, studies on the relationship between perceived human emotions and urban scenes have tended to focus only on the physical characteristics of space. This has also resulted in weak geographical replication of related studies (Kedron et al., 2020).

To close the gap in this effort, thus, using vision language description, we explore participants' emotional changes in city scenes and propose a natural language processing-based framework for the scoring of human perceptions in this study.

## 2    Related Works

In urban research, measuring urban perception has long been a critical part. Physiological (Valtchanov and Ellard, 2015) and psychological (Quercia et al., 2014) measurement techniques have been frequently used in the study. These urban spatial images with geographic location information and physical space settings have been used extensively in recent years due to the rapid growth of map services (such as Google Map) and crowdsourcing. By identifying urban features and detecting urban spatial aspects, these data improve our understanding of the physical space and dynamic qualities of cities. MIT Media Lab introduced the "Place Pulse" project in 2013 (Dubey et al., 2016). And numerous research on how people perceive urban looks have been influenced by the data collection that includes plenty of urban street view images (SVIs) and the corresponding emotional annotations.

Previous approaches, such as semantic segmentation (Ramírez et al., 2021) and object detection (Wang et al., 2022a), however, concentrated primarily on shallow information of SVIs, leading to poor geographic reproducibility. It is also challenging to capture deep semantic information, such as descriptions of urban scenes in human language. With the advancement of computer vision, this issue is resolved via an image caption network. The vision language of the image is acquired using recurrent neural network (RNN) and dynamic target detection. To extract deeper semantic features from SVIs and translate them to visual language, we use an image translation network built on long short-term memory (LSTM) and an attention algorithm with a high SOTA in this study (Anderson et al., 2018).

Natural language processing (NLP) has been widely used in urban studies, such as sentiment analysis of social media (Kong et al., 2022) and interpretation of textual addresses (Qian et al., 2020). It is clear from previous studies that were employing human requires a lot of time and labour while also increasing the error rate. Urban studies employing language models based on RNN and CNN are frequently used due to the rapid development of deep learning. Polysemous word expressions in text present a challenge for these learnt words embedding algorithms. Pre-trained word embedding, a frequently utilized idea in EMLo (Peters et al., 2018), GPT (Radford et al., 2018), and BERT (Devlin et al., 2019), improves this issue. Mainly, BERT combines the advantage of EMLo, and GPT and employs the two-way transformer encoder as the language model framework. In addition, it uses the fine-tuning language model and downstream tasks as the training target, saving a significant amount of training time and improving word understanding. And they have been extremely successful at NLP tasks. In this study, we use a fine-tuned pre-trained BERT model to scoring sentiment in the language of urban scenes.

## 3   Methodology

As shown in Fig. 1, we divided this study into three essential steps. The first step is to transform the deep semantic features of each SVIs in the Place Pulse 2.0 dataset into natural language using the image caption model. Step 2: A fine-tuned pre-trained BERT model was run as a regression network, and we took each street view image's description and associated emotional label from the Place Pulse dataset to create our urban vision language training dataset, which consists of six human perceptual dimensions. Step 3: A text topic algorithm was used to investigate semantic information in order to better understand urban settings and their significance. Moreover, the performance of five traditional methods was compared, and a migrant experiment in Hongkong was conducted.

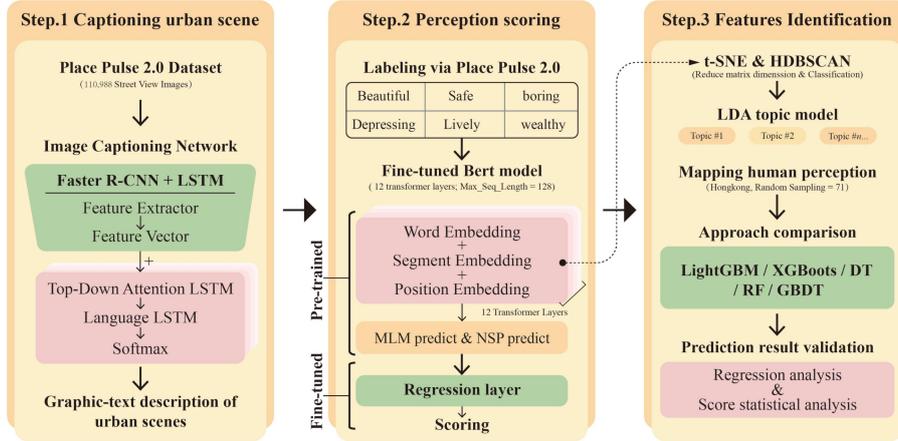

**Fig. 1.** Research framework.

### 3.1   MIT Place Pulse Dataset

The MIT Media Lab launched Place Pulse 2.0, a platform for data collecting, in 2013 to gather online human impression of metropolitan appearance. The participants scored using randomly chosen city street view images from the website, and the dataset includes six perceptual dimensions, namely, beautiful, safety, wealthy, depressing, boring, and lively. The question was, "Which place looks x?" given the two pictures, and the platform's user interface is depicted in Fig. 2. A total of 1,169,078 scoring results were gathered by the end of October 2016, when there were 81,630 online participants (Dubey et al., 2016).

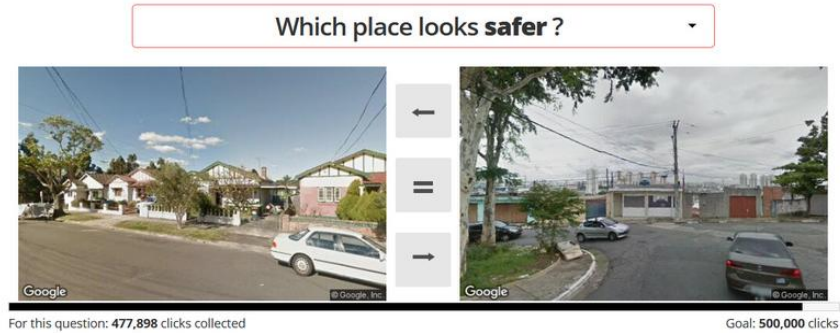

**Fig. 2.** The user interface of the MIT Place Pulse data collection platform.

The dataset includes 110,988 street view images from 56 cities in 28 nations/regions on 6 continents that were captured between 2007 and 2012. Geographical coordinates and camera angles are included in the image metadata. The Place Pulse data scoring approach compares two images, although using a single label sample is more feasible for the application. Using the method proposed by (Zhang et al., 2018), we determined the score for each image sample.

### 3.2   Deep Semantic Feature for Image Caption

Image caption takes images as input, and through mathematical models and calculations, the computer outputs natural language descriptions corresponding to the images (Anderson et al., 2018). The image captioning network can be spread into two parts, encoder and decoder, respectively (Fig. 3). The first part focuses on image feature encoding, which processes through the Faster R-CNN network to extract the image feature to vector. Furthermore, in the second part, an LSTM with top-down attention decoding each feature's global weight and input to a language LSTM network during caption generation to describe the urban scene and generate multiple graphic-text.

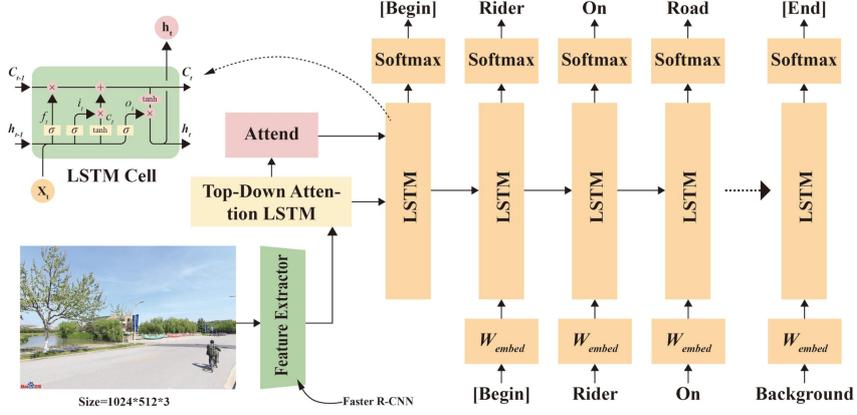

**Fig. 3.** Overview of the image captioning network.

The pre-trained image caption model provided by Luo et al. (2018) in our study and training on COCO dataset that contains more than 1.5milion image description annotations on 330,000 images. In the training process, the model inputs the embedding sequence ($x_1, x_2, ..., x_N$), computes the hidden sequence ($h_1, h_2, ..., h_N$), and obtains the output sequence ($y_1, y_2, ..., y_N$). We add [begin] and [end] to indicate the beginning and end of description process, and the decoder model outputs [end] at $t$ step. The computation procedure is described by the following equation (1):

$$x_{-1} = CNN(image), x_t = Embedding_x, t \in \{0, ..., N-1\},$$
$$h_t = LSTM(x_t, h_{t-1}), t \in \{0, ..., N-1\}, \quad (1)$$
$$y_t = p(y_t|y_{t-1}) = softmax(W_p h_t + b_p). \#$$

where $y$ is final output words, $W_p$ are the embedding weight need training, and $b_p$ are the biases weights. Finally, the image caption model shown a great performance (CIDEr=1.158; SPICE=0.2114).

### 3.3 Fine-tuning the Pre-trained BERT Model

BERT is a natural language processing model based on the transformer convolutional neural network released by Google in 2018, which consists of 12 transformer layers, 12 attention heads and 110M parameters (Devlin et al., 2019). And the pre-trained BERT model saves many computing resources and time compared to training from scratch to achieve a SOTA results. The training process adopts an unsupervised training method and uses token embedding, segment embedding, and position embedding as the input layer to train large-scale corpus. By jointly training the Masked Language Model (MLM) task and the Next Sentence Prediction (NSP) task, the BERT enables the vector representation of each word output as comprehensively, providing better initial parameters for subsequent fine-tuning tasks (Fig. 4a).

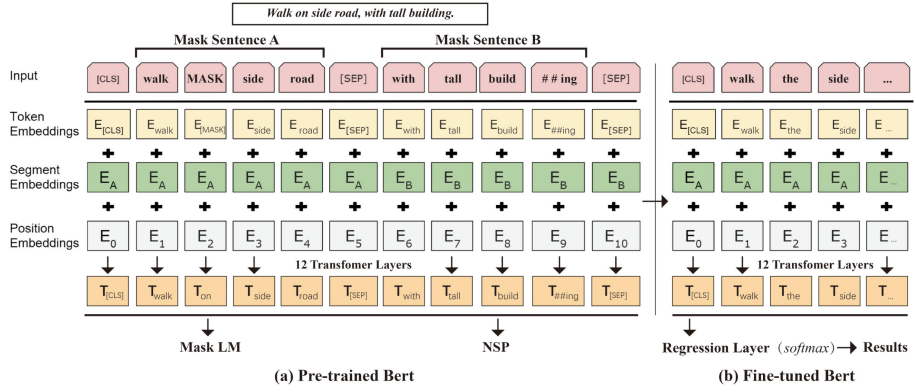

**Fig. 4.** Overview of pre-trained BERT and fine-tuned BERT.

In this study, a new function layer was added after the last transformer encoder layer to scoring the six human perceptions in urban scene (Fig. 4b). Specifically, we "Freeze" most of the pre-trained parameters of the BERT model and update the remaining parameters by adding regression layer as a new function layer and computes the averaged token score. The score calculation equation is given by following (2):

$$Pr\left(l=1|w_{i,j}\right) = softmax(\mathbf{H} \cdot \mathbf{W}_S + \mathbf{b}_s)$$
$$score(s_i) = \frac{\sum_{\forall j} Pr\left(l=1|w_{i,j}\right)}{|s_i|} \quad (2)$$

where $s_i$ is a sequence of tokens $[w_{i,j}]_{j=1}^{|s_i|}$, $x$ is represented by BERT embedding layers $x = ([CLS], s_1, [SEP], ...s_n, [SEP])$, and BERT model hidden representation $\mathbf{H} \in \mathbb{R}^{h \times |L|}$, then hidden layer is passed to a summarization layer $\mathbf{W}_S \in \mathbb{R}^{h \times |L|}$ and computes the token score for each token. $score(s_i)$ is averaged score for each perception.

## 4 Results

### 4.1 Vision Language Representation of the City Scene

Cities often provide street view images that look similar, but the visual descriptions of various viewpoints and locations work together to create the scene features. We perform scene inference on the 1,169,078 images in the place pulse dataset using the visual language model described in Section 3.2. The algorithm creates five of the most pertinent Street View descriptions for each image. On a Tesla P100 GPU, we produced 5,845,390 city-specific text descriptions after 96 hours of inference. We chose at random street view images and the most relevant image description sentences (Fig. 5). This benefit encompasses not only the identification of fundamental elements but also the relationships between entities in terms of position and distance.

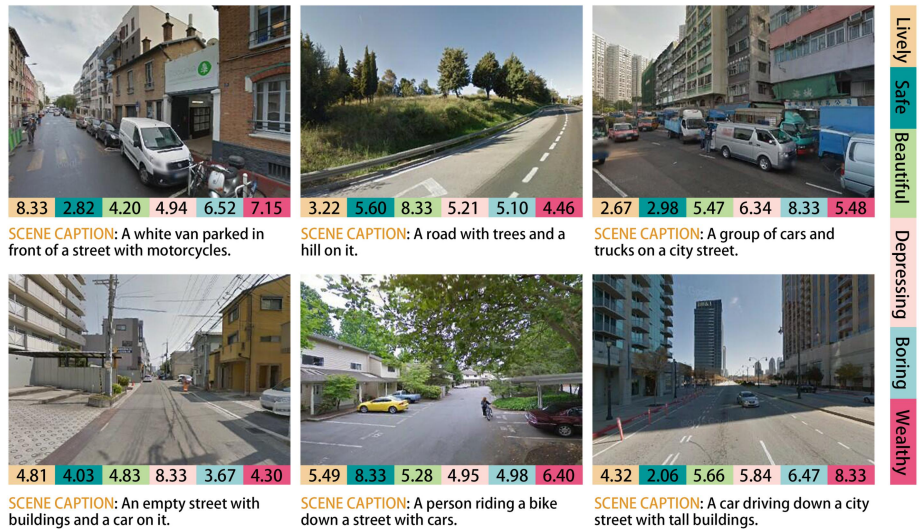

**Fig. 5.** Example of scene descriptions for each perception.

### 4.2 Identify the Urban Scenes Topic

In order to better comprehend the intricate details of urban scene descriptions, we extracted feature matrices (110,988 × 768) from scene-aware descriptions produced by the sentence embedding method in the BERT model. Additionally, we use t-SNE (Van der Maaten and Hinton, 2008) to reduce the dimensionality of the data. 9 separate scenes in the city are discovered by an unsupervised HDBSCAN algorithm clustering (McInnes et al., 2017). Furthermore, we employ a Latent Dirichlet Allocation (LDA) topic model (Peinelt et al., 2020) to extract keywords in various contexts in order to reveal the relationships between the entities in each category (the number of topics is consistent with the number of HDBSCAN clusters, *Topic* = 9).

We discovered that the topic mining results contained certain positional relations, such as on the side, in front, etc., and appeared to be related to scene composition, with features like roads, buildings, vehicles, etc. (Table 1).

**Table 1.** The 5 words with highest probability among the subject words of each urban topic.

| ID   | Topic1     | Topic2   | Topic3   | Topic4     | Topic5    | Topic6     | Topic7  | Topic8    | Topic9   |
|------|------------|----------|----------|------------|-----------|------------|---------|-----------|----------|
| Word | "building" | "road"   | "side"   | "street"   | "street"  | "street"   | "park"  | "drive"   | "car"    |
| Prob | 0.20       | 0.28     | 0.24     | 0.20       | 0.31      | 0.21       | 0.23    | 0.21      | 0.23     |
| Word | "front"    | "tree"   | "park"   | "car"      | "empty"   | "person"   | "car"   | "down"    | "street" |
| Prob | 0.12       | 0.18     | 0.22     | 0.15       | 0.21      | 0.12       | 0.23    | 0.21      | 0.20     |
| Word | "street"   | "empty"  | "car"    | "down"     | "building"| "down"     | "side"  | "car"     | "down"   |
| Prob | 0.12       | 0.08     | 0.20     | 0.13       | 0.15      | 0.10       | 0.17    | 0.19      | 0.18     |
| Word | "house"    | "fence"  | "street" | "city"     | "tree"    | "building" | "group" | "road"    | "drive"  |
| Prob | 0.10       | 0.07     | 0.14     | 0.11       | 0.12      | 0.09       | 0.09    | 0.10      | 0.17     |
| Word | "park"     | "park"   | "two"    | "building" | "house"   | "ride"     | "city"  | "highway" | "group"  |
| Prob | 0.06       | 0.04     | 0.04     | 0.06       | 0.10      | 0.07       | 0.02    | 0.05      | 0.12     |

Topic 3 and Topic 7 have the significant association ($p = 0.9353$), as seen in Figure 6a, which appears to be connected to the fact that both subjects discuss scenarios involving street parking. Various item characteristics and spatial correlations also appear to influence how humans perceive emotions at the same time. We discovered that Topic 2 and Topic 7 showed the highest sensitivity to these perceptions ($p > 0.115$), with Topic 6 having the lowest mean chance of including all six emotional sensations (Fig. 6b).

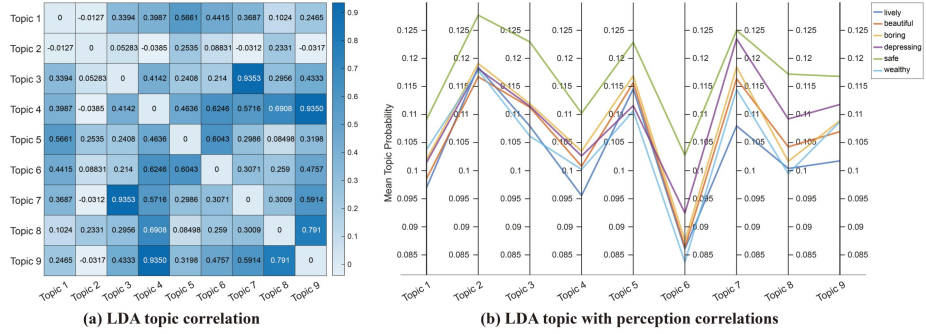

**Fig. 6.** LDA topic correlation (a) and LDA topic with perception correlations (b).

### 4.3 Results Comparison and Migration Experiment

We apply the Fine-tuned BERT model for training the urban scene perception measurement task. The parameter settings are as follows: Batch_Size=32, Learning_Rate=2e-5, and Max_Seq_Length=128. Test dataset is 10%, while the training dataset is 90%. Model Bert_uncased_L-12_H-768_A-12 utilizes as the pre-training model. The experimental platform makes use of TensorFlow 2.1. The two main evaluations that we employ are Mean Square Error (MSE) and $R^2$. The square of the expected difference between the projected value and the actual value is known as the MSE; The model's accuracy increases as the value decreases. $R^2$ measures how interpretable each model variable is in relation to the independent variable regression; the closer the result is to 1, the more accurate the model is.

The results are shown in Table 3. When compared to LightGBM, XGBoots, Decision Tree, Random Forest, and GBDT (the five methods use the same parameters and the training dataset to test dataset ratio is 9:1), Fine-tuned BERT performs best in MAE and $R^2$. When compared to the best baseline approach, our model's accuracy is improved by about 42% overall (Table 2).

**Table 2.** Comparison results of Fine-tuned BERT with the baseline algorithm.

| Model | Lively | | Beautiful | | Boring | | Depressing | | Safe | | wealthy | |
|---|---|---|---|---|---|---|---|---|---|---|---|---|
| | MSE | $R^2$ | MSE | $R^2$ | MSE | $R^2$ | MSE | $R^2$ | MSE | $R^2$ | MSE | $R^2$ |
| **Fine-tuned BERT** | **0.47** | **0.51** | **0.43** | **0.40** | **0.53** | **0.53** | **0.58** | **0.48** | **0.58** | **0.25** | **0.55** | **0.45** |
| LightGBM | 0.92 | 0.05 | 0.80 | 0.05 | 1.05 | 0.04 | 1.18 | 0.05 | 1.18 | 0.05 | 1.10 | 0.05 |
| XGBoost | 0.77 | 0.20 | 0.68 | 0.18 | 0.91 | 0.17 | 1.01 | 0.18 | 1.01 | 0.19 | 0.93 | 0.19 |

| | | | | | | | | | | | |
|---|---|---|---|---|---|---|---|---|---|---|---|
| Decision Tree | 0.96 | 0.07 | 0.83 | 0.08 | 1.09 | 0.01 | 1.23 | 0.01 | 1.24 | 0.01 | 1.15 | 0.01 |
| Random Forest | 0.95 | 0.04 | 0.83 | 0.01 | 1.10 | 0.02 | 1.24 | 0.01 | 1.23 | 0.01 | 1.14 | 0.02 |
| GBDT | 0.89 | 0.08 | 0.77 | 0.08 | 1.02 | 0.07 | 1.15 | 0.07 | 1.15 | 0.08 | 1.07 | 0.07 |

We also conduct migration experiments in Hong Kong using the model. After considering the scene text, we applied the refined BERT model to predict the perception scores of each dimension at 71 randomly generated locations. In parallel, to validate the model results, we also gathered 15 residents to rate each of the six perception variables. The findings demonstrate the highly consistency($R^2 = 0.42$) between the predictions made using Fine-tuned BERT and the results of manual scoring(Fig. 7), while also showing the viability of utilizing natural language techniques to assess the human perception of urban space (Fig. 8).

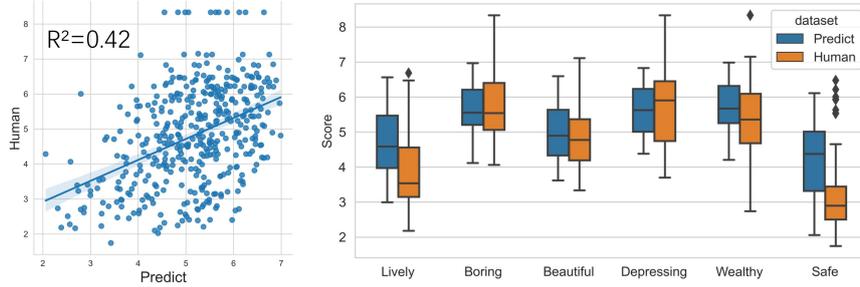

**Fig. 7.** Relationship between predicted and manual scoring.

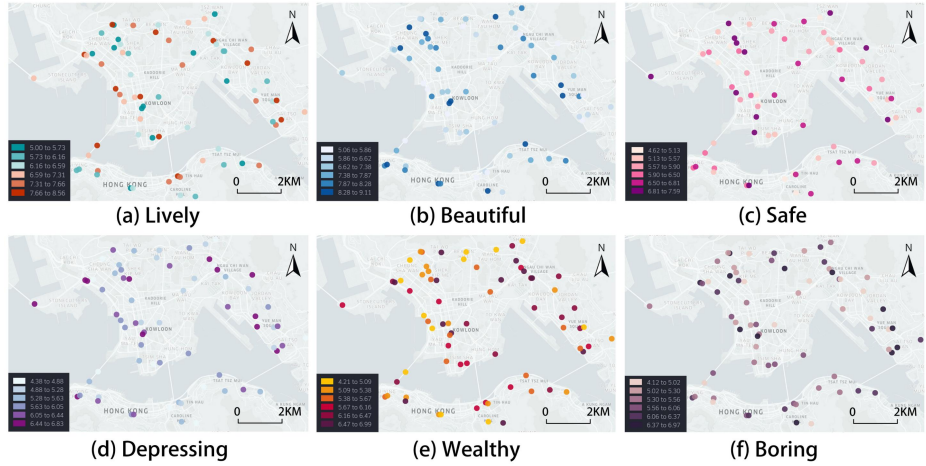

**Fig. 8.** Migration experiment results in Hongkong

## 5  Conclusion and Discussion

This study introduces a measurement framework for perceiving understandable urban scenes based on a visual language model and a pre-trained BERT model. Use the 110,988 street view images in the Place Pulse 2.0 dataset as regression data, including the scene descriptions and related emotional scores. The results demonstrate that (i) the image caption model is more accurate for the text description of street view images, and these texts contain the entity attributes and location attributes of scene objects; (ii) the approach of combining BERT embedding vector and LDA topic model is beneficial to mining urban scenes deep semantic characteristics of the nine categories of urban scenes, all of which have distinct category features; and (iii) compared with traditional measuring methods (such as machine learning), the six perceptual dimension measurement tasks based on the pre-trained BERT model have a high SOTA.

Of course, there are some limitations on this study. The first is that our vision language model is limited by its inability to recognize the intangible urban content, which limits its ability to distinguish scenes without visual differences. The second is that urban scenes are constantly changing and that street view images can only capture urban scenes at a particular moment. An expected future work direction is to further expand the analysis of the causes of urban perception phenomena by combining richer multivariate urban data.